\documentclass[englishn,runningheads]{llncs}
\usepackage[T1]{fontenc}
\usepackage[latin1]{inputenc} 
\usepackage{times}
\usepackage{graphicx}
\usepackage{epsfig}
\usepackage{epic}  
\usepackage{float}
\usepackage{color}
\usepackage[english]{babel}
\usepackage[shortend,boxed,vlined,english]{algorithm2e}
\usepackage{subfigure}
\usepackage{psfrag}
\usepackage{wrapfig}
\usepackage{gastex}
\usepackage{pspictpg}

\usepackage{amsmath}    
\usepackage{amssymb}    

\makeatletter
\usepackage{babel}
\makeatother

\begin{document}

\title{Using Pseudo-Stochastic Rational Languages in Probabilistic Grammatical Inference\thanks{This work was partially supported by the Marmota project ANR-05-MMSA-0016}}

\author{Amaury Habrard \and François Denis \and Yann Esposito}

\institute{Laboratoire d'Informatique Fondamentale de Marseille (L.I.F.) UMR
CNRS 6166
\email{\{habrard,fdenis,esposito\}@cmi.univ-mrs.fr}}

\maketitle

\begin{abstract}

 In probabilistic grammatical inference, a usual goal is to infer a
good approximation of an unknown distribution $P$ called a {\it
stochastic language}. The estimate of $P$ stands in some class of
probabilistic models such as probabilistic automata (PA). In this
paper, we focus on probabilistic models based on multiplicity automata
(MA). The stochastic languages generated by MA are called
\emph{rational stochastic languages}; they strictly include
stochastic languages generated by PA; they also admit a very concise
canonical representation. Despite the fact that this class is not
recursively enumerable, it is efficiently identifiable in the limit by
using the algorithm DEES, introduced by the authors in a previous
paper.  However, the identification is not proper and before the
convergence of the algorithm, DEES can produce MA that do not define
stochastic languages. Nevertheless,  it is possible to use these MA to define
stochastic languages. We show that they belong to a broader class
of rational series, that we call \emph{pseudo-stochastic rational
languages}.  The aim of this paper is twofold. First we provide a
theoretical study of pseudo-stochastic rational languages, the
languages output by DEES, showing for example that this class is
decidable within polynomial time. Second, we have carried out a lot of
experiments in order to compare DEES to classical inference
algorithms such as ALERGIA and MDI. They show that DEES outperforms
them in most cases.
 
\end{abstract}

\section{Introduction}

In probabilistic grammatical inference, we often consider stochastic
languages which define distributions over $\Sigma^*$, the set of all
the possible words over an alphabet $\Sigma$. In general, we consider an unknown
distribution $P$  and the goal is to find a good approximation
given a finite sample of words independently drawn from $P$. 

The class of probabilistic automata (PA) is often used for modeling
such distributions. This class has the same expressiveness as Hidden
Markov Models and is identifiable in the
limit~\cite{DenisEsposito04}. However, there exists no efficient
algorithm for identifying PA. This can be explained by the fact that
there exists no canonical representation of these automata which makes
it difficult to correctly identify the structure of the target. One solution is to
focus on subclasses of PA such as probabilistic deterministic
automata~\cite{COC94,TDH00} but with an important lack of
expressiveness. Another solution consists in considering the class of
multiplicity automata (MA). These models admit a canonical
representation which offers good opportunities from a machine learning
point of view.  MA define functions that compute rational series with
values in ${\mathbb R}$~\cite{DenisEsposito06}. MA are a strict
generalization of PA and the stochastic languages generated by PA are
special cases of rational stochastic languages. Let us denote by
$S^{rat}_K(\Sigma)$ the class of rational stochastic languages
computed by MA with parameters in $K$ where
$K\in\{\mathbb{Q},\mathbb{Q}^+,\mathbb{R},\mathbb{R}^+\}$. With
$K=\mathbb{Q}^+$ or $K=\mathbb{R}^+$, $S^{rat}_K(\Sigma)$ is exactly
the class of stochastic languages generated by PA with parameters in
$K$. But, when $K=\mathbb{Q}$ or $K=\mathbb{R}$, we obtain strictly
greater classes. This provides several advantages: Elements of
$S^{rat}_K(\Sigma)$ have a minimal normal representation, thus
elements of ${\cal S}_{K^+}^{rat}(\Sigma)$ may have significantly
smaller representation in ${\cal S}_{K}^{rat}(\Sigma)$; parameters of
these minimal representations are directly related to probabilities of
some natural events of the form $u\Sigma^*$, which can be efficiently
estimated from stochastic samples; lastly when $K$ is a field,
rational series over $K$ form a vector space and efficient linear
algebra techniques can be used to deal with rational stochastic
languages.

However, the class ${\cal S}_{{\mathbb Q}}^{rat}(\Sigma)$ presents a
  serious drawback: There exists no recursively enumerable subset
  class of MA which exactly generates it~\cite{DenisEsposito04}. As a
  consequence, no proper identification algorithm can exist: indeed,
  applying a proper identification algorithm to an enumeration of
  samples of $\Sigma^*$ would provide an enumeration of the class of
  rational stochastic languages over ${\mathbb Q}$. In spite of this
  result, there exists an efficient algorithm, DEES, which is able to
  identify $S^{rat}_K(\Sigma)$ in the limit. But before reaching the
  target, DEES can produce MA that do not define stochastic
  languages. However, it has been shown in~\cite{DEH_colt06} that with
  probability one, for any rational stochastic language $p$, if DEES
  is given as input a sufficiently large sample $S$ drawn according to
  $p$, DEES outputs a rational series such that $\sum_{u\in
  \Sigma^*}r(u)$ converges absolutely to 1. Moreover, $\sum_{u\in
  \Sigma^*}|p(u)-r(u)|$ converges to 0 as the size of $S$
  increases. We show that these MA belong to a broader class of
  rational series, that we call \emph{pseudo-stochastic rational
  languages}. A pseudo-stochastic rational language $r$ has the
  property that $r(u\Sigma^*)=lim_{n\rightarrow \infty}r(u\Sigma^{\leq
  n})$ is defined for any word $u$ and that $r(\Sigma^*)=1$. A
  stochastic language $p_r$ can be associated with $r$ in such a way
  that $\sum_{u\in \Sigma^*}|p_r(u)-r(u)|=2\sum_{r(u)<0}|r(u)|$ when
  the sum $\sum_{u\in \Sigma^*}r(u)$ is absolutely convergent. As a
  first consequence, $p_r=r$ when $r$ is a stochastic language. As a
  second consequence, for any rational stochastic language $p$, if
  DEES is given as input increasing samples drawn according to $p$,
  DEES outputs pseudo-stochastic rational languages $r$ such that $\sum_{u\in
  \Sigma^*}|p(u)-p_r(u)|$ converges to 0 as the size of $S$
  increases.

The aim of this paper is twofold: To provide a theoretical study of the
  class of pseudo-stochastic rational languages and a series of
  experiments in order to compare the performance of DEES to two classical
  inference algorithms: ALERGIA~\cite{COC94} and MDI~\cite{TDH00}.
We show that the class of pseudo-stochastic rational languages is
decidable within polynomial time. We provide an algorithm that can be
used to compute $p_r(u)$ from any MA that computes $r$. We also show
how it is possible to simulate $p_r$ using such an automaton. We show
that there exist pseudo-stochastic rational languages $r$ such that
$p_r$ is not rational. Finally, we show that it is undecidable whether
two pseudo-stochastic rational languages define the same stochastic
language.
We have carried out a lot of experiments which show that DEES
  outperforms ALERGIA and MDI in most cases. These results were
  expected since ALERGIA and MDI have not the same theoretical
  expressiveness and since DEES aims at producing a minimal
  representation of the target in the set of MA, which can be
  significantly smaller than the smaller equivalent PDA (if it
  exists).

The paper is organized as follows. In section 2, we introduce some
background about multiplicity automata, rational series and stochastic
languages and present the algorithm DEES. Section 3 deals with our
study of pseudo-rational stochastic languages. Our experiments are
detailed in Section 4.

\section{Definitions and notations}

\subsection{Rational series, multiplicity automata and stochastic languages}

Let $\Sigma^{*}$ be the set of words on the finite alphabet $\Sigma$.
A language is a subset of $\Sigma^{*}$.  The empty word is denoted by
$\varepsilon$ and the length of a word $u$ is denoted by $|u|$.  For
any integer $k$, let $\Sigma^k=\{u\in \Sigma^{*}: \ |u|=k\}$ and
$\Sigma^{\leq k}=\{u\in \Sigma^{*}: \ |u|\leq k\}$. We denote by $<$
the length-lexicographic order on $\Sigma^*$ and by $Min L$ the
minimal element of a non empty language $L$ according to this order. A
subset $S$ of $\Sigma^{*}$ is \emph{prefix-closed} if for any $u,v\in
\Sigma^*$, $uv\in S\Rightarrow u\in S$. For any $S\subseteq \Sigma^*$,
let $pref(S)=\{u\in \Sigma^*: \exists v\in \Sigma^*, uv\in S\}$ and
$fact(S)=\{v\in \Sigma^*: \exists u,w\in \Sigma^*, uvw\in S\}$.

A \emph{formal power series} is a mapping $r$ of $\Sigma^*$ into
  ${\mathbb R}$.  The set of all formal power series is denoted by
  ${\mathbb R}\langle\langle \Sigma\rangle\rangle$. It is a vector
  space. For any series $r$ and any word $u$, let us denote by
  $\dot{u}r$ the series defined by $\dot{u}r(w)=r(uw)$ for every word
  $w$.  Let us denote by $supp(r)$ the \emph{support} of $r$, i.e. the
  set $\{w\in \Sigma^*: r(w)\neq 0\}$.
A \emph{stochastic language} is a formal series $p$ which takes its
values in ${\mathbb R}^+$ and such that $\sum_{w\in \Sigma^*}p(w)=1$.
 The set of all stochastic languages over $\Sigma$
is denoted by ${\cal S}(\Sigma)$. 
For any language $L\subseteq \Sigma^*$ and any $p\in {\cal S}(\Sigma)$, let us denote $\sum_{w\in
L}p(w)$ by $p(L)$. 
For any $p\in {\cal S}(\Sigma)$ and $u\in\Sigma$ such that
  $p(u\Sigma^*)\neq 0$, the \emph{residual language} of $p$ wrt $u$ is
  the stochastic language defined by $u^{-1}p$ by
  $u^{-1}p(w)=\frac{p(uw)}{p(u\Sigma^*)}$.
We denote by $res(p)$ the set $\{u\in \Sigma^*: p(u\Sigma^*)\neq 0\}$
and by $Res(p)$ the set $\{u^{-1}p: u\in res(p)\}$.

Let $S$ be a sample over $\Sigma^*$, i.e. a multiset composed of words
over $\Sigma^*$. We denote by $p_{S}$ the empirical distribution
over $\Sigma^*$ associated with $S$. Let $S$ be an infinite sample
composed of words independently drawn according to a stochastic
language $p$. We denote by $S_n$ the sequence composed of the $n$
first words of $S$.

We introduce now the notion of multiplicity automata (MA).
Let $K\in \{{\mathbb R},{\mathbb Q},
{\mathbb R}^+, {\mathbb Q}^+ \}$. A $K$-\emph{multiplicity automaton (MA)} is a
  5-tuple $\langle \Sigma,Q,$ $\varphi, \iota,\tau\rangle $ where
 $Q$
  is a finite set of states,
 $\varphi:Q\times\Sigma\times Q\rightarrow
  K$ is the transition function,
 $\iota:Q\rightarrow K$ is the
  initialization function,
 $\tau:Q\rightarrow K$ is the termination
  function.
We extend
  the transition function $\varphi$ to $Q\times\Sigma^*\times Q$ by
  $\varphi(q,wx,r)=\sum_{s\in Q}\varphi(q,w,s)$ $\varphi(s,x,r)$ and
  $\varphi(q,\varepsilon,r)=1$ if $q=r$ and $0$ otherwise, for any
  $q,r\in Q$, $x\in \Sigma$ and $w\in \Sigma^*$. For any finite subset
  $L\subset \Sigma^*$ and any $R\subseteq Q$, define
  $\varphi(q,L,R)=\sum_{w\in L, r\in R}\varphi(q,w,r)$.
We denote by  $Q_I=\{q\in Q|\iota(q)\neq 0\}$  the set of
  \emph{initial states} and by $Q_T=\{q\in Q|\tau(q)\neq 0\}$ the set
  of \emph{terminal states}.
A state $q\in Q$ is \emph{accessible}
(resp. \emph{co-accessible}) if there exists $q_0\in Q_I$
(resp. $q_t\in Q_T$) and $u\in \Sigma^*$ such that
$\varphi(q_0,u,q)\neq 0$ (resp. $\varphi(q,u,q_t)\neq 0$).  An MA is
\emph{trimmed} if all its states are accessible and
co-accessible. 
From now, we only consider trimmed MA.
The \emph{support} of an MA
  $A=\left\langle \Sigma,Q,\varphi,\iota,\tau\right\rangle$ is the
  \emph{Non-deterministic Finite Automaton (NFA)}
  $\langle \Sigma,Q,Q_I, Q_T, \delta\rangle$ where
  $\delta(q,x)=\{q'\in Q|\varphi(q,x,q')\neq 0\}$.

The \emph{spectral radius} of a square matrix $M$ if the maximum
        magnitude of its eigenvalues. Let $A=\langle \Sigma, Q=\{q_1,
        \ldots, q_n\}, \iota, \varphi, \tau\rangle$ be an MA. Let us
        denote by $\rho(A)$ be the spectral radius of the square
        matrix $[\varphi(q_i,\Sigma,q_j)]_{1\leq i,j\leq n}$
        ($\rho(A)$ does not depends on the order of the states). If
        $\rho(A)<1$ then each sequence $r_{A,q}(\Sigma^{\leq n})$
        converges to a number $s_q$ and hence, $r(\Sigma^{\leq n})$
        converges too~\cite{DEH_colt06}. Let us denote by $r(\Sigma^*)$ the
limit of $r(\Sigma^{\leq n})$ when it exists. 
The numbers $s_q$ are the unique
        solutions of the following linear system of equations (and
        therefore are computable within polynomial time):\\[2pt]
\centerline{$s_q=r_{A,q}+\sum_{q'\in Q}\varphi(q,\Sigma,q')s_{q'}\textrm{ for
}q\in Q.$}
It is decidable within polynomial time whether
$\rho(A)<1$~\cite{BlondelTsitsiklis00,Gantmacher66}.

A \emph{Probabilistic Automaton (PA)} is a trimmed MA $\left\langle
\Sigma,Q,\varphi,\iota,\tau\right\rangle $ s.t. $\iota, \varphi$ and $\tau$
take their values in $[0,1]$, s.t. $\sum_{q\in Q}\iota(q)=1$ and
for any state $q$, $\tau(q)+\varphi(q,\Sigma,Q)=1$.  A \emph{Probabilistic Deterministic Automaton (PDA)} is a \emph{PA} whose
support is deterministic. 
It can be shown that Probabilistic Automata generate stochastic
languages.  Let us denote by ${\cal S}_K^{PA}(\Sigma)$ (resp. ${\cal
S}_K^{PDA}(\Sigma)$) the class of all stochastic languages which can
be computed by a $PA$ (resp. a $PDA$).

For any MA $A$,
let $r_A$ be the series defined by $r_A(w)=\sum_{q, r\in
Q}\iota(q)$ $\varphi(q,w,r) \tau(r)$. 
\noindent For any $q\in Q$, we also define the series $r_{A,q}$ by
$r_{A,q}(w)=\sum_{r\in Q}\varphi(q,w,r)\tau(r)$.  An MA $A$ is
\emph{reduced} if the set $\{r_{A,q}| q\in Q\}$ is linearly independent
in the vector space ${\mathbb
R}\langle\langle\Sigma\rangle\rangle$. An MA $A$ is
\emph{prefix-closed} if (i) its set of states $Q$ is a prefix-closed
subset of $\Sigma^*$, (ii) $Q_I=\{\varepsilon\}$ and (iii) $\forall
u\in Q, \delta(\varepsilon,u)=\{u\}$ where $\delta$ is the transition
function in the support of $A$.

Rational series have several
characterization~(\cite{BerstelReutenauer84,SalomaaSoittola78}). Here,
we shall say that a formal power series over $\Sigma$ is $K$-rational
iff there exists a $K$-multiplicity automaton $A$ such that $r=r_A$,
where $K\in \{{\mathbb R}, {\mathbb R^+}, {\mathbb Q}, {\mathbb
Q^+}\}$.  Let us denote by $K^{rat}\langle\langle
\Sigma\rangle\rangle$ the set of $K$-rational series over $\Sigma$ and
by ${\cal
S}_K^{rat}(\Sigma)=K^{rat}\langle\langle\Sigma\rangle\rangle\cap {\cal
S}(\Sigma)$, the set of \emph{rational stochastic languages} over
$K$. It can be shown that a series $r$ is ${\mathbb R}$-rational iff
the set $\{\dot{u}r|u\in \Sigma^*\}$ spans a finite dimensional vector
subspace of ${\mathbb R}\langle\langle\Sigma\rangle\rangle$. As a
corollary, a stochastic language $p$ is ${\mathbb R}$-rational iff the
set $Res(p)$ spans a finite dimensional vector subspace $[Res(p)]$ of ${\mathbb
R}\langle\langle\Sigma\rangle\rangle$.  Rational stochastic languages
have been studied in \cite{DenisEsposito06} from a language
theoretical point of view. It is worth noting that ${\cal
S}^{PDA}_{\mathbb R}(\Sigma)\subsetneq {\cal S}^{PA}_{\mathbb
R}(\Sigma)= {\cal S}^{rat}_{\mathbb R^+}(\Sigma)\subsetneq {\cal
S}^{rat}_{\mathbb R}(\Sigma)$. From now on, a rational stochastic
language will always denote an ${\mathbb R}$-rational stochastic
language.

Rational stochastic languages have a serious drawback. There exists no
recursively enumerable subset of multiplicity automata capable to
generate them~\cite{DenisEsposito04,DenisEsposito06}. As a consequence, it is
undecidable whether a given MA computes a stochastic language.

Every rational language is the support of a rational series but the
converse is false: there exists rational series whose supports are not
rational. For example, it can be shown that the complementary set of
$\{a^nb^n | n \in {\mathbb N}\}$ in $\{a,b\}^*$ is the support of a
rational series. However, a variant of Pumping Lemma holds for
languages which are support of rational series. Let $L$ be such a
language. There exists an integer $N$ such that for any word $w=uv\in
L$ satisfying $|v|\geq N$, there exists $v_1,v_2, v_3$ such that
$v=v_1v_2v_3$ and $L\cap uv_1v_2^*v_3$ is infinite~\cite{BerstelReutenauer84}.

Rational stochastic languages admit a canonical representation by
reduced prefix-closed MA. 
Let $p$ be a rational
stochastic language and let $Q_p$ be the smallest basis of $[Res(p)]$
(for the order induced by $<$ on the finite subsets of $\Sigma^*$).
Let $A=\left\langle \Sigma,Q_p,\varphi,\iota,\tau\right\rangle $ be
the MA defined by: {\it (i)}
 $\iota(\varepsilon)=1$, $\iota(u)=0$ otherwise;
 $\tau(u)=u^{-1}p(\varepsilon)$, 
{\it (ii)} $\varphi(u,x,ux)=u^{-1}p(x\Sigma^*)$ if $u,ux\in Q_p$ and
$x\in \Sigma$, 
{\it (iii)} $\varphi(u,x,v)=\alpha_vu^{-1}p(x\Sigma^*)$ if $x\in
\Sigma$, $ux\in (Q_p\Sigma\setminus Q_p) \cap res(p)$ and
$(ux)^{-1}p=\sum_{v\in Q_p}\alpha_vv^{-1}p$. 
It can be shown that $A$ is a reduced prefix-closed MA which computes
$p$ and such that $\rho(A)<1$. $A$ is called the \emph{canonical
representation} of $p$. Note that the parameters of $A$ correspond to
natural components of the residual of $p$ and can be estimated by
using samples of $p$.

\subsection{Inference of rational stochastic languages}
The algorithm DEES~\cite{DEH_colt06} is able to identify rational
stochastic languages: with probability one, for every rational
stochastic language $p$ and every infinite sample $S$ of $p$, there
exists an integer $N$ such that for every $n\geq N$, DEES($S_n$)
outputs the canonical representation $A$ of $p$.  Before its
presentation, we introduce informally the basic idea of the algorithm.
First, the goal is to find the structure of the automaton, {\it i.e.}
the set of states $Q_p$ smallest basis of $[Res(P)]$.  The inference
proceeds as follows: the algorithm begins by building a unique state
which corresponds to the residual $\epsilon^{-1}p_S$. Each state of
the automaton corresponds to some residual $u^{-1}p_s$ where $u$ is
the prefix of some examples in $S$. After having built a state
corresponding to $u^{-1}p_s$, for any letter $x$, the algorithm
studies the possibility of adding a new state corresponding to
$(ux)^{-1}p_s$ or of creating transitions labeled by $x$ that lead to
the states already built in the automaton.  A new state will be added
to the automaton if the residual language corresponding to
$(ux)^{-1}p_s$ cannot be approximated as a linear combination of the
residual languages corresponding the states already built.

The pseudo-code of the algorithm is presented in Algorithm~\ref{A:DEES}.
In order to find a linear combination, DEES uses the following set of
inequalities where $S$ is a non empty finite sample of $\Sigma^*$,  $Q$ 
a prefix-closed subset of $pref(S)$,  $v\in pref(S)\setminus Q$, and 
$\epsilon>0$:\\
\centerline{\small $I(Q,
v,S,\epsilon)=\{|v^{-1}P_S(w\Sigma^*)-\sum_{u\in
Q}X_uu^{-1}P_S(w\Sigma^*)|\leq \epsilon | w\in
fact(S)\}\cup\{\sum_{u\in Q}X_u=1\}.$}\\

\begin{algorithm}[t]
\KwIn{a sample $S$}
\KwOut{a prefix-closed reduced
MA $A=\left\langle \Sigma,Q,\varphi,\iota,\tau\right\rangle $}

$Q\leftarrow\left\{ \varepsilon\right\} $;\hspace{.5cm}
$\iota(\varepsilon)\leftarrow 1$ ; \hspace{.5cm}
$\tau(\varepsilon)\leftarrow P_S(\varepsilon)$\;
$F\leftarrow\Sigma\cap pref(S)$ /*F is the frontier set*/\;

\While{$F\neq\emptyset$}{
$v\leftarrow Min F$ s.t. $v=u.x$ where $u\in \Sigma^*$
  and $x\in \Sigma$\; $F\leftarrow F\setminus\left\{ v\right\}
  $\;
\eIf{$I(Q, v,S,|S|^{-1/3})$ has no solution}{
$Q\leftarrow Q\cup\left\{ v\right\} $; \hspace{.5cm} $\iota(v)\leftarrow 0$; \hspace{.5cm} $\tau(v)\leftarrow P_S(v)/P_S(v\Sigma^*)$\;

$\varphi(u,x,v)\leftarrow P_S(v\Sigma^*)/P_S(u\Sigma^*)$;\hspace{.5cm}
$F\leftarrow
F\cup\{vx\in res(P_S)|x\in\Sigma$\}\}\;
}{
\SetLine let $(\alpha_w)_{w\in Q}$ be a solution of $I(Q, v,S,|S|^{-1/3})$\; 
\lForEach{$w\in Q$}{$\varphi(u,x,w)\leftarrow \alpha_wP_S(v\Sigma^*)/P_S(u\Sigma^*)$\;}
}
}

\caption{Algorithm DEES.\label{A:DEES}}
\end{algorithm}


DEES runs in polynomial time in the size of $S$ and identifies in the
limit the structure of the canonical representation $A$ of the target
$p$. Once the correct structure of $A$ is found, the algorithm
computes estimates $\alpha_S$ of each parameter $\alpha$ of $A$ such
that $|\alpha-\alpha_S|= O(|S|^{-1/3})$. The output automaton $A$
computes a rational series $r_A$ such that $\sum_{w\in
\Sigma^*}r_A(w)$ converges absolutely to 1. Moreover, it can be shown
that $r_A$ converges to the target $p$ under the $D1$ distance (also
called the $L1$ norm), stronger than distance $D_2$ or $D_{\infty}$:
$\sum_{w\in \Sigma^*}|r_A(w)-p(w)|$ tends to 0 when the size of $S$
tends to $\infty$. If the parameters of $A$ are rational numbers, a
variant of DEES can identify exactly the target~\cite{DEH_colt06}.

\begin{figure}[tb]
\unitlength=1cm
\gasset{Nw=0.75,Nh=0.75,Nmr=0.5}
\begin{center}
\begin{tabular}{ccc}
\subfigure[Initialisation with $\varepsilon$.\label{EXDEES1}]{
\begin{picture}(3.5,2)(0,0)\nullfont
\node[Nmarks=i](0)(1.5,1.5){$\varepsilon$}
\node[Nframe=n,Nfill=n](0a)(1.5,0.1){}
\drawedge[ELside=r](0,0a){$\frac{1}{6}$}
\end{picture}
}

&
\subfigure[Creation of a new state.\label{EXDEES2}]{
\hspace{.25cm}
\begin{picture}(4,2)(0,0)\nullfont      
\node[Nmarks=i](0)(0.5,1.5){$\varepsilon$}
\node[Nframe=n,Nfill=n](0a)(0.5,0.1){}
\node[Nmarks=n](1)(2.5,1.5){$a$}
 \node[Nframe=n,Nfill=n](1a)(2.5,0.1){}
\drawedge[ELside=r](0,0a){$\frac{1}{6}$}
\drawedge[ELside=r](0,1){$$}
\drawedge[ELside=r](1,1a){$\frac{2}{5}$}
\end{picture}
}
&
\subfigure[Final automaton.\label{EXDEES3}]{
\hspace{.5cm}
\begin{picture}(4,2)(0,0)\nullfont
      \gasset{Nw=0.75,Nh=0.75,Nmr=0.5}
\node[Nmarks=i](0)(0.5,1.5){$\varepsilon$}
\node[Nframe=n,Nfill=n](0a)(0.5,0.1){}
\node(1)(2.5,1.5){$a$}
\node[Nframe=n,Nfill=n](1a)(2.5,0.1){}
\drawedge[ELside=r](0,0a){$\frac{1}{6}$}
 \drawedge[ELside=r](0,1){$a, \frac{5}{6}$}
\drawedge[ELside=r](1,1a){$\frac{2}{5}$}
\gasset{curvedepth=-0.45}\drawedge[ELside=r](1,0){$a, -\frac{3}{10}$}
\drawloop[loopangle=90](1){$a, \frac{9}{10}$}
\end{picture}
}
\end{tabular}
\end{center}
\vspace{-1cm}
\caption{Illustration of the different steps of algorithm DEES.}\vspace{-.6cm}
\end{figure}

\noindent We give now a simple example that illustrates DEES.  Let us consider a
sample $S=\{\varepsilon,a,aa,aaa\}$ such that $|\varepsilon|=10$, $|a|=|aa|=20$,
 $|aaa|=10$. We have the following values for the empirical
distribution: $P_S(\varepsilon)=P_S(aaa)=P_S(aaa\Sigma^*)=\frac{1}{6}$, $P_S(a)=P_S(aa)=\frac{1}{3}$, 
$P_S(a\Sigma^*)=\frac{5}{6}$, $P_S(aa\Sigma^*)=\frac{1}{2}$ and $P_S(aaaa\Sigma^*)=0$,
$\varepsilon=\frac{1}{(60)^{\frac{1}{3}}}\equiv 0.255$. With the sample $S$,
DEES will infer a multiplicity automaton in three steps:
\begin{enumerate}
\item We begin by constructing a state for $\varepsilon$  (Figure~\ref{EXDEES1}). 
\item We examine $P_S(v\Sigma^*)$ with $v=\varepsilon a$ to decide if we need to add a new state for the string $a$.
We obtain the following system which has in fact no solution and we create a new state as shown in Figure~\ref{EXDEES2}.
$${\footnotesize\begin{array}{lcl}
\left\{\left| \frac{P_S(va\Sigma^*)}{P_S(v\Sigma^*)} -
\frac{P_S(a\Sigma^*)}{P_S(\Sigma^*)}* X_\varepsilon \right|  \leq
b\right.,&\hspace{1cm} &
\left| \frac{P_S(vaa\Sigma^*)}{P_S(v\Sigma^*)} -
\frac{P_S(aa\Sigma^*)}{P_S(\Sigma^*)}* X_\varepsilon \right|  \leq b,\\[.25cm]
\left| \frac{P_S(vaaa\Sigma^*)}{P_S(v\Sigma^*)} -
  \frac{P_S(aaa\Sigma^*)}{P_S(\Sigma^*)}* X_\varepsilon \right|  \leq b,& &
\left.
X_\varepsilon = 1\phantom{ \frac{P_S(va\Sigma^*)}{P_S(v\Sigma^*)}}\right\}
\end{array}}
$$
\item 
We examine $P_S(v\Sigma^*)$ with $v=aa$ to decide if we need to create a new state for the string $aa$.
We obtain the system below. It is easy to see that this system admits
at least one solution $X_\varepsilon =-\frac{1}{2}$ and
$X_a=\frac{3}{2}$. Then, we add two transitions to the automaton and we obtain the automaton of Figure~\ref{EXDEES3} and the
algorithm halts.
$$
{\footnotesize
\begin{array}{ll}
\hspace{-.75cm}\left\{ \left| \frac{P_S(va\Sigma^*)}{P_S(v\Sigma^*)} -
\frac{P_S(a\Sigma^*)}{P_S(\Sigma^*)} X_\varepsilon
-\frac{P_S(aa\Sigma^*)}{P_S(a\Sigma^*)} X_a \right|  \leq b,\ \right.& \hspace{-1.4cm}
\left| \frac{P_S(vaa\Sigma^*)}{P_S(v\Sigma^*)} -
\frac{P_S(aa\Sigma^*)}{P_S(\Sigma^*)} X_\varepsilon -
\frac{P_S(aa\Sigma^*)}{P_S(a\Sigma^*)} X_a \right|  \leq b,\\
\left| \frac{P_S(vaaa\Sigma^*)}{P_S(v\Sigma^*)} -
\frac{P_S(aaa\Sigma^*)}{P_S(\Sigma^*)}  X_\varepsilon -
\frac{P_S(aaa\Sigma^*)}{P_S(a\Sigma^*)} X_a \right|  \leq b, & \hspace{1cm}
\left.X_\varepsilon + X_a = 1\phantom{\frac{P_S(va\Sigma^*)}{P_S(v\Sigma^*)}}\right\}
\end{array}}
$$
\end{enumerate}

Since no recursively enumerable subset of MA is capable to generate
the set of rational stochastic languages, no identification algorithm
can be proper. This remark applies to DEES.  There is no guarantee at
any step that the automaton $A$ output by DEES computes a stochastic
language. However, the rational series $r$ computed by the MA output
by DEES can be used to compute a stochastic language $p_r$ that also
converges to the target~\cite{DEH_colt06}. Moreover, they have several nice
properties which make them close to stochastic languages: We call them
pseudo-stochastic rational languages and we study their properties in
the next Section.
\section{Pseudo-stochastic rational languages}
The canonical representation $A$ of a rational stochastic language
satisfies $\rho(A)<1$ and $\sum_{w\in \Sigma^*}r_A(w)=1$. We use this
characteristic to define the notion of pseudo-stochastic rational
language.

\begin{definition}
We say that a rational
series $r$ is a \emph{pseudo-stochastic language} if there exists an
MA $A$ which computes $r$ and such that $\rho(A)<1$ and if $r(\Sigma^*)=1$.
\end{definition}
Note that the condition $\rho(A)<1$ implies that $r(\Sigma^*)$ is
defined without ambiguity. A rational stochastic language is a
pseudo-stochastic rational language but the converse is false.

\textbf{Example.}  Let $A=\left\langle
  \Sigma,\{q_0\},\varphi,\iota,\tau\right\rangle $ defined by
$\Sigma=\{a,b\}$, $\iota(q_0)=\tau(q_0)=1$, $\varphi(q_0,a,q_0)=1$ and
$\varphi(q_0,b,q_0)=-1$. We have $r_A(u)=(-1)^{|u|_b}$. Check that
$\rho(A)=0$ and $r_A(u\Sigma^*)=(-1)^{|u|_b}$ for every word
$u$. Hence, $r_A$ is a pseudo stochastic language.

As indicated in the previous section, any canonical representation $A$ of a rational stochastic language
satisfies $\rho(A)<1$. In fact, the next Lemma shows that any reduced
representation $A$ of a pseudo-stochastic language satisfies
$\rho(A)<1$.

\begin{lemma}\label{lemmaRHO}
Let $A$ be a reduced representation of a
pseudo-stochastic language. Then, $\rho(A)<1$.
\end{lemma}

\begin{proof}
The proof is detailed in Annex~\ref{AnnexlemmaRHO}.
\end{proof}

\begin{proposition}
  It is decidable within polynomial time whether a given MA computes a
  pseudo-stochastic language.
\end{proposition}
\begin{proof}
Given an MA $B$, compute a reduced representation $A$ of $B$, check
whether $\rho(A)<1$ and then, compute $r_A(\Sigma^*)$. \qed
\end{proof}

It has been shown in~\cite{DEH_colt06} that a stochastic language $p_r$ can
be associated with a pseudo-stochastic rational language $r$: the idea
is to prune in $\Sigma^*$ all subsets $u\Sigma^*$ such that
$r(u\Sigma^*)\leq 0$ and to normalize in order to obtain a stochastic
language. Let $N$ be the smallest prefix-closed subset of $\Sigma^*$
satisfying\\
\centerline{$\varepsilon\in N\textrm{ and }\forall u\in N, x\in \Sigma, ux\in N
\textrm{\ iff\ }r(ux\Sigma^*)>0.$}\\[.01pt]

For every $u\in \Sigma^*\setminus N$,
define $p_r(u)=0$. For every $u\in N$, let
$\lambda_u=Max(r(u),0)+\sum_{x\in \Sigma}Max(r(ux\Sigma^*),0)$. Then,
define $p_r(u)=Max(r(u),0)/\lambda_u$. It can be shown
(see~\cite{DEH_colt06}) that $r(u)\leq 0 \Rightarrow p_r(u)=0\textrm{ and
}r(u)\geq 0 \Rightarrow r(u)\geq p_r(u).$ 

\noindent The difference between $r$
and $p_r$ is simple to express when the sum $\sum_{u\in \Sigma^*}r(u)$
converges absolutely. Let $N_r=\sum_{r(u)\leq 0}|r(u)|$. We have
$\sum_{w\in\Sigma^*}|r(u)-p_r(u)|=N_r+\sum_{r(u)>0}(r(u)-p_r(u))=2N_r+\sum_{u\in
\Sigma^*}(r(u)-p_r(u))=2N_r.$ Note that when $r$ is a stochastic
language,$\sum_{u\in \Sigma^*}r(u)$ converges absolutely and
$N_r=0$. As a consequence, in that case, $p_r=r$. 
We give in Algorithm~\ref{A:PR} an algorithm that computes $p_r(u)$ and $p_r(u\Sigma^*)$
for any word $u$ from any MA that computes $r$.
 This algorithm is linear in the length of the input. It can
 be slightly modified to generate a word drawn according to $p_r$ (see
 Annex~\ref{APR}).

\begin{algorithm}[t]
\KwIn{MA
    $A=\left\langle \Sigma,Q=\{q_1,
    \ldots,q_n\},\varphi,\iota,\tau\right\rangle $ s.t. $\rho(A)<1$ and
     $r_A(\Sigma^*)=1$\\ a word $u$ }
\KwOut{$p_{r_A}(u), p_{r_A}(u\Sigma^*)$}
\For{$i=1,\ldots,n$ /* this step is polynomial in $n$ and is done
  once*/}{
$s_i\leftarrow r_{A,q_i}(\Sigma^*)$; \hspace{.5cm} $e_i\leftarrow \iota(q_i)$\;
}
$w\leftarrow \varepsilon$; \hspace{.5cm} $\lambda \leftarrow 1$ /* $\lambda$ is
equal to $p_{r_A}(w\Sigma^*)$*/ \;
\Repeat{$w=u$}{
 $\mu\leftarrow \sum_{i=1}^ne_i\tau(q_i)$;\hspace{.5cm} 
$S\leftarrow \{(w,Max(\mu,0))\}$\; 
\For{$x\in \Sigma$}{
$\mu\leftarrow \sum_{i,j=1}^ne_i\varphi(q_i,x,q_j)s_j$;\hspace{.5cm} 
$S\leftarrow S\cup \{(wx,Max(\mu,0))\}$\; }
\SetLine $\sigma\leftarrow\sum_{(v,\mu)\in S}\mu$;\hspace{.5cm}
$S\leftarrow\{(x,\mu/\sigma)|(x,\mu)\in S\}$ /*normalization*/ \;
 \lIf{$w=u$}{$p_{r_A}(u)\leftarrow \lambda\mu$ /*where $(u,\mu)\in S$ and
  $\lambda=p_{r_A}(u\Sigma^*)$*/\;}
\Else{{\bf Let} $x\in \Sigma$ s.t. $wx$ is a prefix of $u$ and let $\mu$
  s.t. $(wx,\mu)\in S$\;
$w\leftarrow wx$; \hspace{.5cm}$\lambda\leftarrow \lambda\mu$; \hspace{.5cm}
\lFor{$i=1,\ldots,n$}{$e_i\leftarrow \sum_{j=1}^ne_j\varphi(q_j,x,q_i)$ \;}
}
}
\caption{Algorithm computing $p_r$.\label{A:PR}}
\end{algorithm}

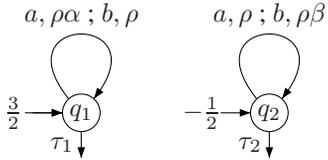
\begin{wrapfigure}{l}{4.5cm}
  \begin{picture}(30,15)(0,0)\nullfont
      \gasset{Nw=5,Nh=5,Nmr=2.5}
\node[Nframe=n,Nfill=n](0b)(0,8.5){$\frac{3}{2}$}
\node[Nmarks=i](0)(9,8.5){$q_1$}
\node[Nframe=n,Nfill=n](0a)(9,0){}
\node(1)(34,8.5){$q_2$}
\node[Nframe=n,Nfill=n](1b)(25,8.5){$-\frac{1}{2}$}
\node[Nframe=n,Nfill=n](1a)(34,0){}
\drawedge(0b,0){}
\gasset{curvedepth=0}

\drawedge[ELside=r](0,0a){$\tau_1$}
\drawloop[loopangle=90](0){$a,\rho\alpha $ ; $b, \rho$}
\gasset{curvedepth=0}
\drawedge(1b,1){}
 
\drawedge[ELside=r](1,1a){$\tau_2$}

\gasset{curvedepth=0}
\drawloop[loopangle=90](1){$a,\rho$ ; $b, \rho\beta$}
\end{picture}
\vspace{-.5cm}
  \caption{An example of pseudo-stochastic rational languages which
    are not
    rational.\label{fig:pseudoratnonrat}}\vspace{-2cm}
\end{wrapfigure}

\noindent The stochastic languages $p_r$ associated with pseudo-stochastic
 rational languages $r$ can be not rational. 
\begin{proposition}
There exists pseudo-stochastic rational languages $r$ such that $p_r$
is not rational.
\end{proposition}
\begin{proof}
Suppose that the parameters of the automaton $A$ described on
  Figure~\ref{fig:pseudoratnonrat} satisfy
  $\rho(\alpha+1)+\tau_1=1$ and $\rho(\beta+1)+\tau_2=1$ with $\alpha>\beta>1$.
  Then the series $r_{q_1}$ and $r_{q_2}$ are rational stochastic
  languages and therefore, $r_A=3r_{q_1}/2-r_{q_2}/2$ is a rational
  series which satisfies $\sum_{u\in \Sigma^*}|r_A(u)|\leq 2$ and
  $\sum_{u\in \Sigma^*}r_A(u)=1$.
  
  \noindent Let us show that $p_{r_A}$ is not rational. For any $u\in \Sigma^*$,
  $r_A(u)=\frac{\rho^{|u|}}{2}(3\alpha^{|u|_a}\tau_1-\beta^{|u|_b}\tau_2).$
  For any integer $n$, there exists an integer $m_n$ such that for any
  integer $i$, $r_A(a^nb^i)>0$ iff $i\leq m_n$. Moreover, it is clear
  that $m_n$ tends to infinity with $m$. Suppose now that $p_{r_A}$ is
  rational and let $L$ be its support. From the Pumping Lemma, there
  exists an integer $N$ such that for any word $w=uv\in L$ satisfying
  $|v|\geq N$, there exists $v_1,v_2, v_3$ such that $v=v_1v_2v_3$ and
  $L\cap uv_1v_2^*v_3$ is infinite. Let $n$ be such that $m_n\geq N$
  and let $u=a^n$ and $v=b^{m_n}$. Since $w=uv\in L$, $L\cap a^nb^*$
  should be infinite, which is is false. Therefore, $L$ is not the
  support of a rational language.\qed 
\end{proof}

Different rational series may yield the same pseudo-rational stochastic
language. Is it decidable whether two pseudo-stochastic rational series define the same
stochastic language? Unfortunately, the answer is no. The proof relies
on the following result: it is undecidable whether a multiplicity
automaton $A$ over $\Sigma$ satisfies $r_A(u)\leq 0$ for every $u\in
\Sigma^*$~\cite{SalomaaSoittola78}. It is easy to show that this result still holds for the set
of MA $A$ which satisfy $|r_A(u)|\leq \lambda^{|u|}$, for any
$\lambda>0$.

\begin{proposition}\label{prop2MA}
It is undecidable whether two rational series define the same
stochastic language.
\end{proposition}

\begin{proof}
 The proof is detailed in Annex~\ref{Annexprop2MA}.
\end{proof}

\section{Experiments}

In this section, we present a set of experiments allowing us to study
the performance of the algorithm DEES for learning good stochastic language
models.  Hence, we will study the behavior of DEES with samples
of distributions generated from PDA, PA and non rational
stochastic language. 
We decide to compare DEES to the most well known probabilistic
grammatical inference approaches: The algorithms
{\it Alergia}~\cite{COC94} and {\it MDI}~\cite{TDH00} that are able to
identify PDAs. These algorithms can be tuned by a parameter, in the
experiments we choose the best parameter which gives the best result
on all the samples, but we didn't change the parameter according to
the size of the sample in order to take into account the impact of the
sample sizes.

In our experiments, we use two performance criteria. We measure the
size of the inferred models by the number of states. Moreover, to
evaluate the quality of the automata, we use the $D1$
norm\footnote{Note that we can't use the Kullback-Leibler measure
 because it is not robust with null probability strings which
 implies to smooth the learned models, and also because automata produced by DEES do not always define stochastic
  language, {\it i.e.} some strings may have a negative value.} between two models $A$ and $A'$ defined by~:\\[1.5pt]
\centerline{$
D1(A,A') = \sum_{u\in \Sigma^*} \left| P_A(u) - P_{A'}(u)\right|.
$\\[2.5pt]}
$D1$ norm is the strongest distance after  Kullback
Leibler. In practice, we use an approximation  by considering a
subset of $\Sigma^*$ generated by $A$ ($A$ will be the target for us).\\

\begin{figure}[t]
          \centering
          \includegraphics[width=11.5cm]{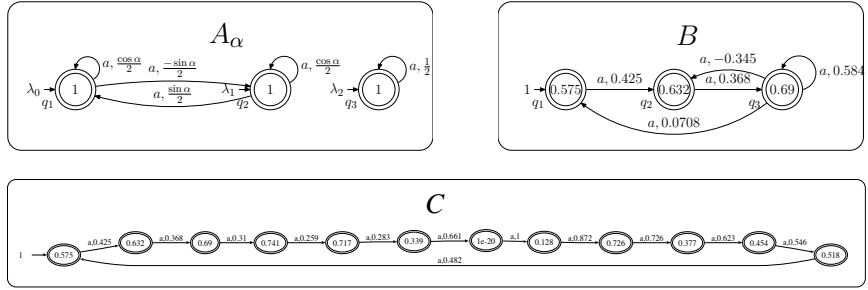}\vspace{-.25cm}
\caption{$A_\alpha$ define stochastic language which can be represented by a PA with at least $2n$ states when $\alpha=\frac{\pi}{n}$. With $\lambda_0=\lambda_2=1$ and $\lambda_1=0$, the MA
$A_{\pi/6}$ defines a stochastic language $P$ whose prefixed reduced
representation is the MA $B$ (with approximate values on
transitions). In fact, $P$ can be computed by a PDA and the smallest PA
computing it is $C$.}\label{fig:FMA}\vspace{-.6cm}
        \end{figure}

\begin{figure}[t]
\begin{center}
\begin{tabular}{ll}
\subfigure[Results with distance $D1$]{\includegraphics[angle=270,width=5.5cm]{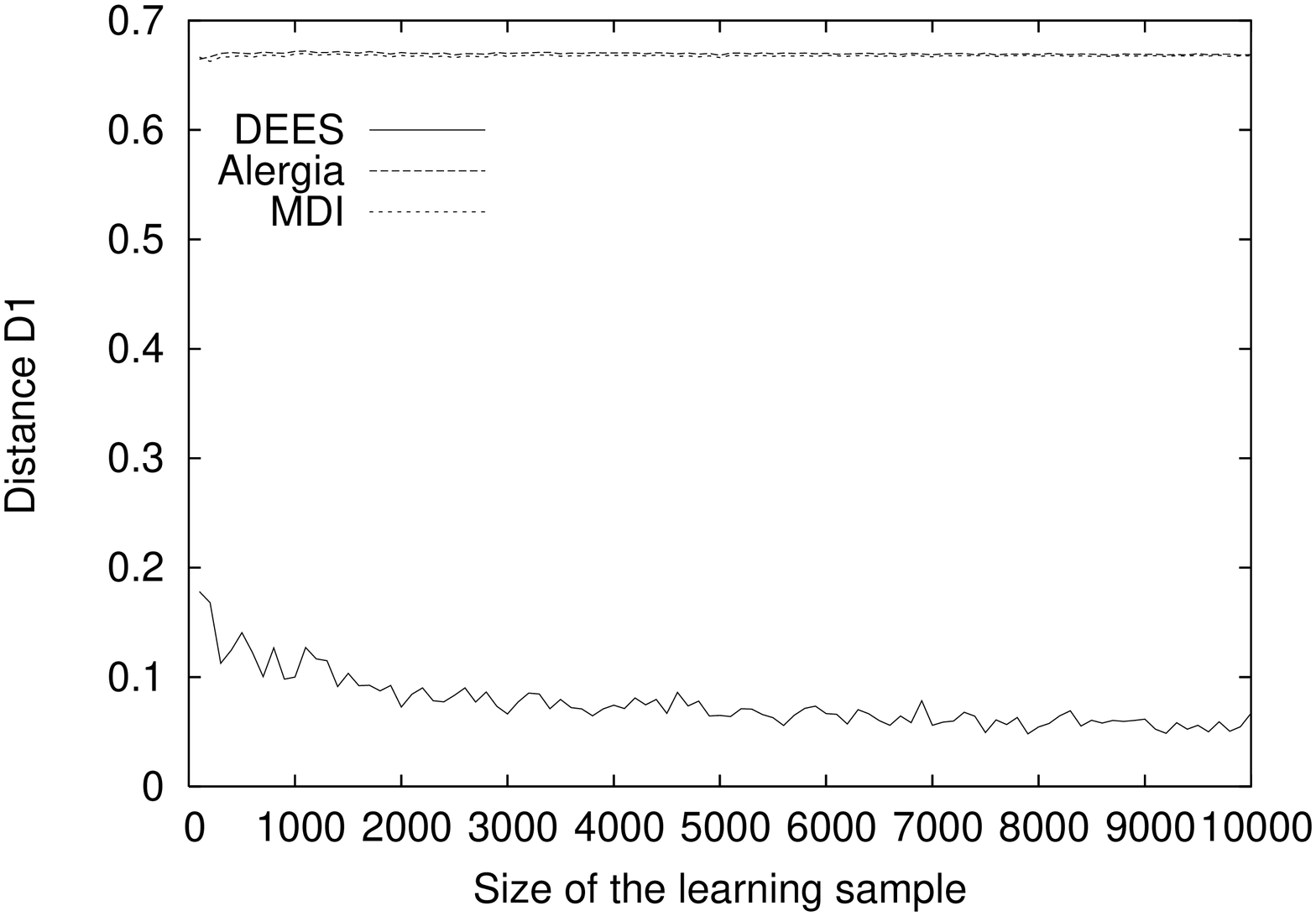}}&
\subfigure[Size of the model.]{\includegraphics[angle=270,width=5.5cm]{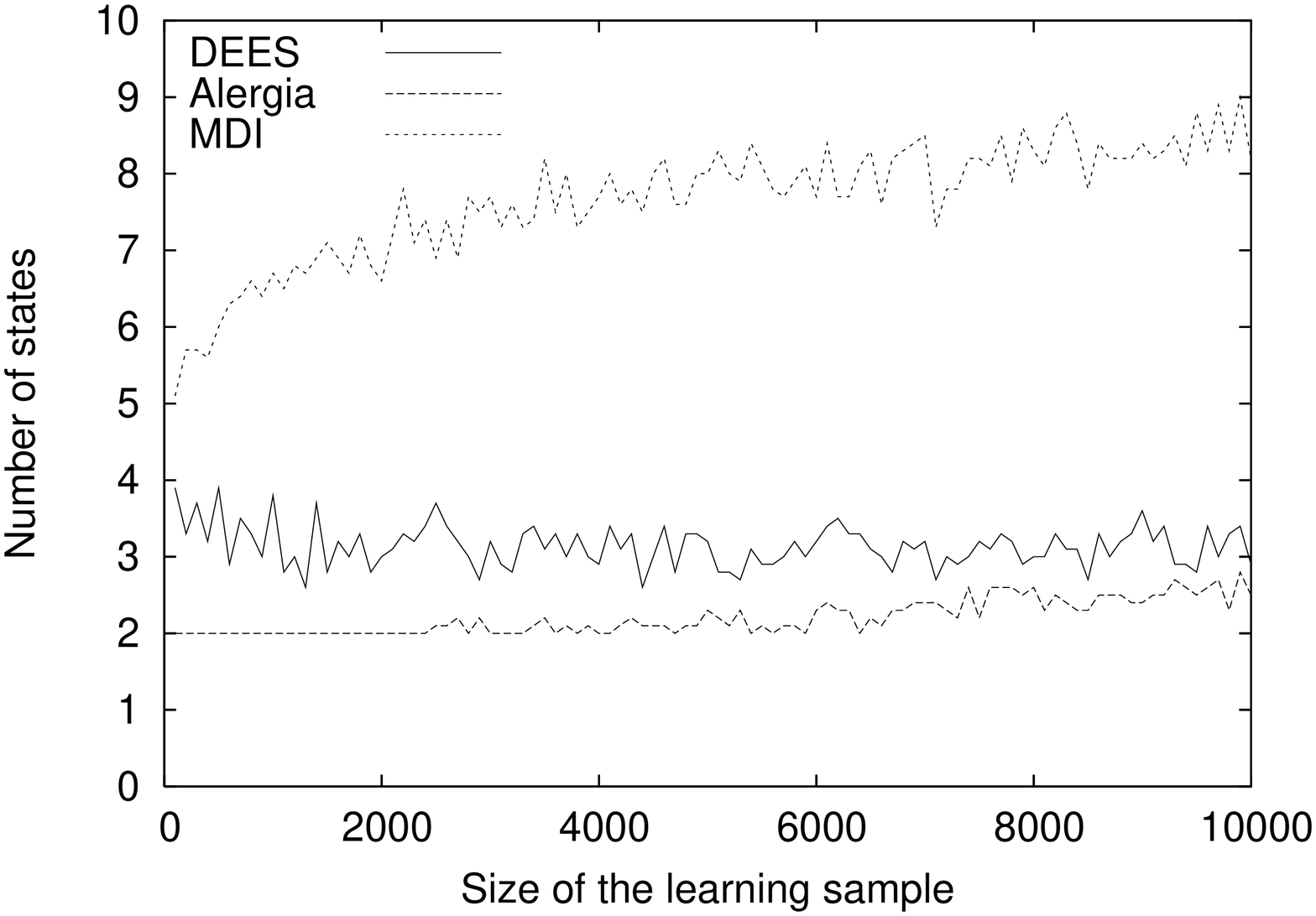}} 
\end{tabular}
\end{center}\vspace{-.9cm}
\caption{Results obtained with the prefix reduced multiplicity
  automaton of three states of Figure~\ref{fig:FMA} admitting a
  representation with a PDA of twelve states. \label{FMA}}\vspace{-.6cm}
\end{figure}

We carried out a first series of experiment where the target automaton
can be represented by a PDA. We consider a stochastic language
defined by the automaton  on
Figure~\ref{fig:FMA}. This stochastic language can be represented by a multiplicity
automaton of three states and by an equivalent minimal PDA of twelve states \cite{DEH_colt06} (Alergia and MDI can then identify this automaton). 
To compare the performances of the three algorithms, we used the
following experimental set up.  From the target automaton, we
generate samples from  size 100 to 10000.
 Then, for each sample we learn an automaton with the three
algorithms and compute the norm $D1$ between them and the
target. We repeat this experimental setup 10 times and give the
average results.
 Figure~\ref{FMA} reports the results obtained. If we consider the size
 of the learned models,  DEES finds quickly
the target automaton, while MDI only begins to tend to the target PDA after 10000 examples. The automata produced by Alergia
are far from this target. This behavior can be explained by the fact
that these two algorithms need
significantly longer examples to find the correct target and thus
larger samples, this is also amplified  because there are more parameters to estimate. In practise we
noticed that the correct structure can be found after more than 100000 examples.
If we look at the distance $D1$, DEES
outperforms MDI and Alergia (which have the same behavior) and begins to converge after 500 examples.

\begin{figure}[t]
\begin{tabular}{cc}
\subfigure{
\hspace{.1cm} $A:$ \hspace{.5cm} 
\begin{picture}(30,15)(0,0)\nullfont
      \gasset{Nw=5,Nh=5,Nmr=2.5}
\node[Nframe=n,Nfill=n](0b)(0,1.5){$\frac{1}{2}$}
\node[Nmarks=i](0)(9,1.5){$1$}
\node[Nframe=n,Nfill=n](0a)(9,-7){}
\node(1)(34,1.5){$2$}
\node[Nframe=n,Nfill=n](1b)(25,1.5){$\frac{1}{2}$}
\node[Nframe=n,Nfill=n](1a)(34,-7){}
\drawedge(0b,0){}
\gasset{curvedepth=0}

\drawedge[ELside=r](0,0a){$\frac{1}{4}$}
\drawloop[loopangle=90](0){$a \frac{\alpha^{2}}{4}$ ; $b, -\frac{\alpha^{-2}}{4}$}
\gasset{curvedepth=0}
\drawedge(1b,1){}
 
\drawedge[ELside=r](1,1a){$\frac{1}{4}$}

\gasset{curvedepth=0}
\drawloop[loopangle=90](1){$a \frac{\alpha^{-2}}{4}$ ; $b, \frac{\alpha^{2}}{4}$}
\end{picture}
\vspace{.25cm}
}
\hspace{2cm}
&
$B:$ \hspace{.75cm}
\subfigure{
\begin{picture}(30,20)(0,0)\nullfont
      \gasset{Nw=5,Nh=5,Nmr=2.5}
\node[Nmarks=i](0)(0,1.5){$\varepsilon$}
\node[Nframe=n,Nfill=n](0a)(0,-7){}
\node(1)(22,1.5){$a$}
\node[Nframe=n,Nfill=n](1a)(22,-7){}
\drawedge[ELside=r](0,0a){$\frac{1}{4}$}
\drawloop[loopangle=90](0){$b, \frac{3}{4}$}
\gasset{curvedepth=3}
 \drawedge(0,1){$a, \frac{3}{8}$;$b, -\frac{3}{8}$}
\gasset{curvedepth=0}
\drawedge[ELside=r](1,1a){$\frac{1}{4}$}
\gasset{curvedepth=3}
\drawedge(1,0){$a, -\frac{1}{6}$ ; $b, \frac{1}{6}$}
\gasset{curvedepth=0}
\drawloop[loopangle=90](1){$a, \frac{3}{4}$}
\end{picture}
}
\vspace{.25cm}

\end{tabular}
\caption{Automaton $A$ is a PA 
  with non rational parameters in  $\mathbb{R}^+$
  ($\alpha=(\sqrt{5}+1)/2$). $A$ can be represented by an MA $B$ with
  rational parameters in $\mathbb{Q}$ \cite{DenisEsposito06}.\label{MARp}}\vspace{-.6cm}
\end{figure}
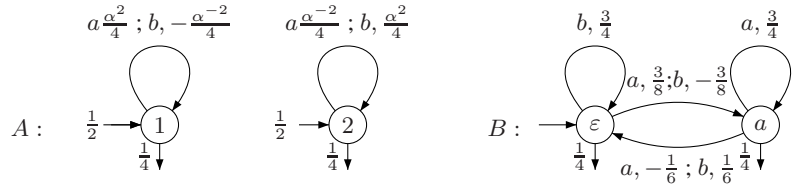

\begin{figure}[t]
\begin{center}
\begin{tabular}{ll}
\subfigure[Results with distance $D1$]{\includegraphics[angle=270,width=5.5cm]{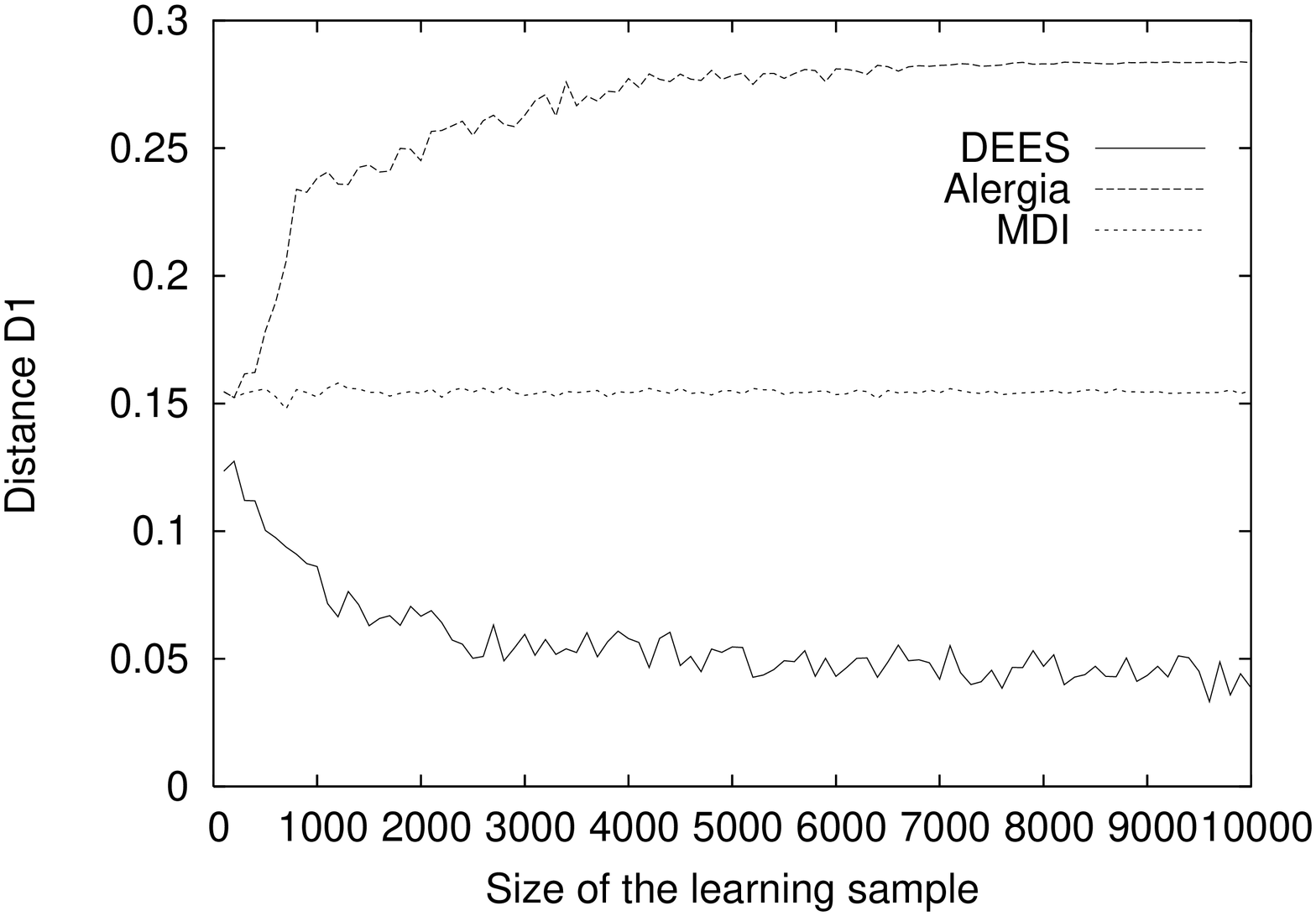}}& 
\subfigure[Size of the model.]{\includegraphics[angle=270,width=5.5cm]{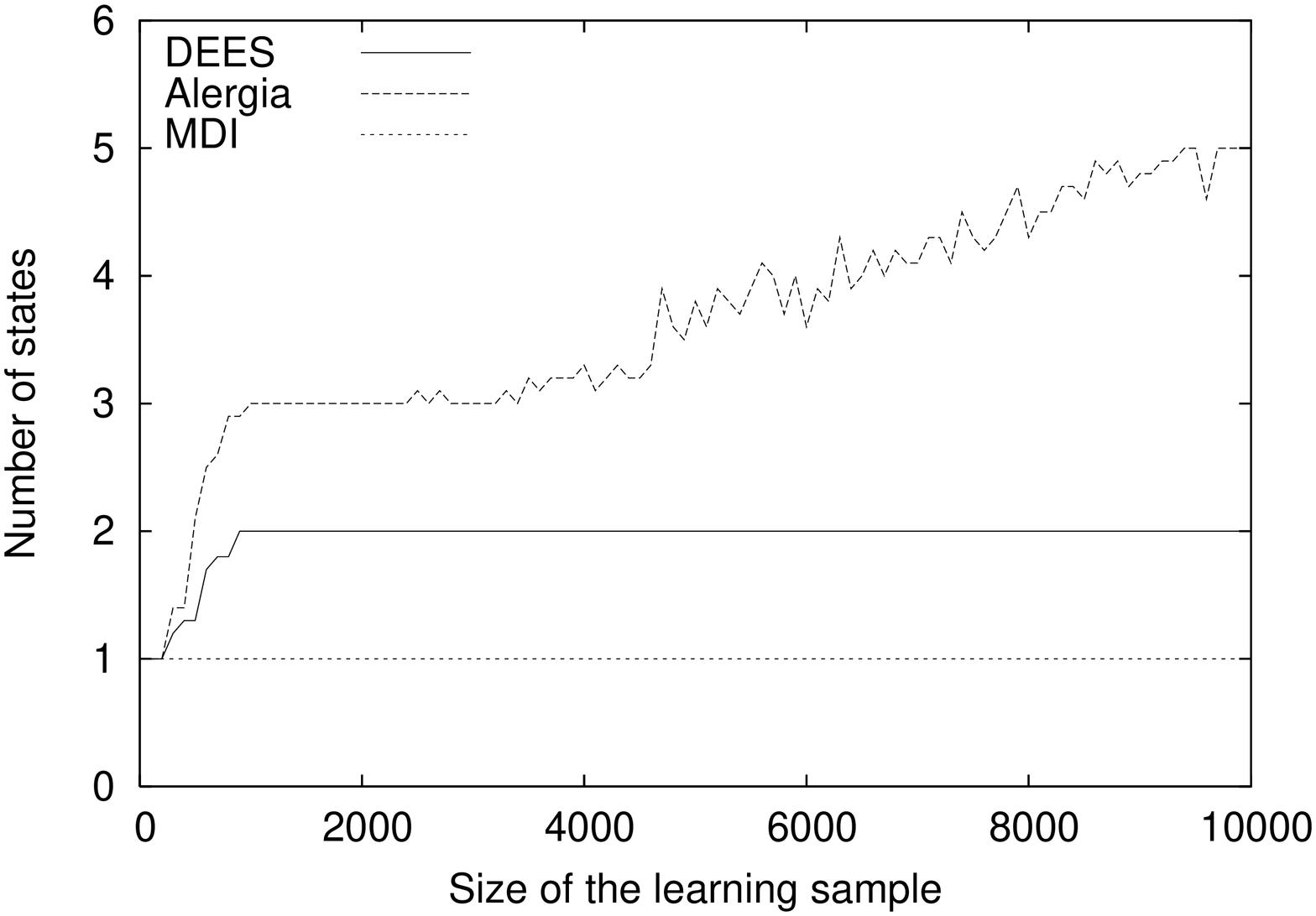}} 
\end{tabular}
\end{center}\vspace{-.9cm}
\caption{Results obtained with the target automaton of
  Figure~\ref{MARp} admitting a representation in the class PA with non rational parameters.\label{ResRp}}\vspace{-.6cm}
\end{figure}

We carried out other series of experiments for evaluating DEES when
the target belongs to the class of PA. First, we consider the simple automaton of Figure~\ref{MARp} which defines a
stochastic language that can be represented by a PA with parameters in
$\mathbb{R}^+$. We follow the same experimental setup as in the first experiment, the results are reported
on Figure~\ref{ResRp}. 
According to our 2 performance criteria, DEES
outperforms again Alergia and MDI. In fact, the target can not be modeled correctly by
Alergia and MDI because it can not be represented by a PDA. This
explains why these algorithms can't find a good model. For them, the
best answer is to produce a unigram model.
 Alergia even diverge at a given
step (this behavior is due to its fusion criterion that becomes more
restrictive with the increasing of the learning set) and   MDI returns
always the unigram.  DEES finds the correct
structure quickly and begins to converge  
after 1000 examples. This behavior confirms the fact DEES can produce
better models with small samples because it constructs small
representations. On the other hand, Alergia and MDI seem to need a
huge number of examples to find a good approximation of the target,
even when the target is relatively small.

\begin{figure}[t]
\begin{center}
\begin{tabular}{ll}
\subfigure[Results with distance $D1$]{\includegraphics[angle=270,width=5.5cm]{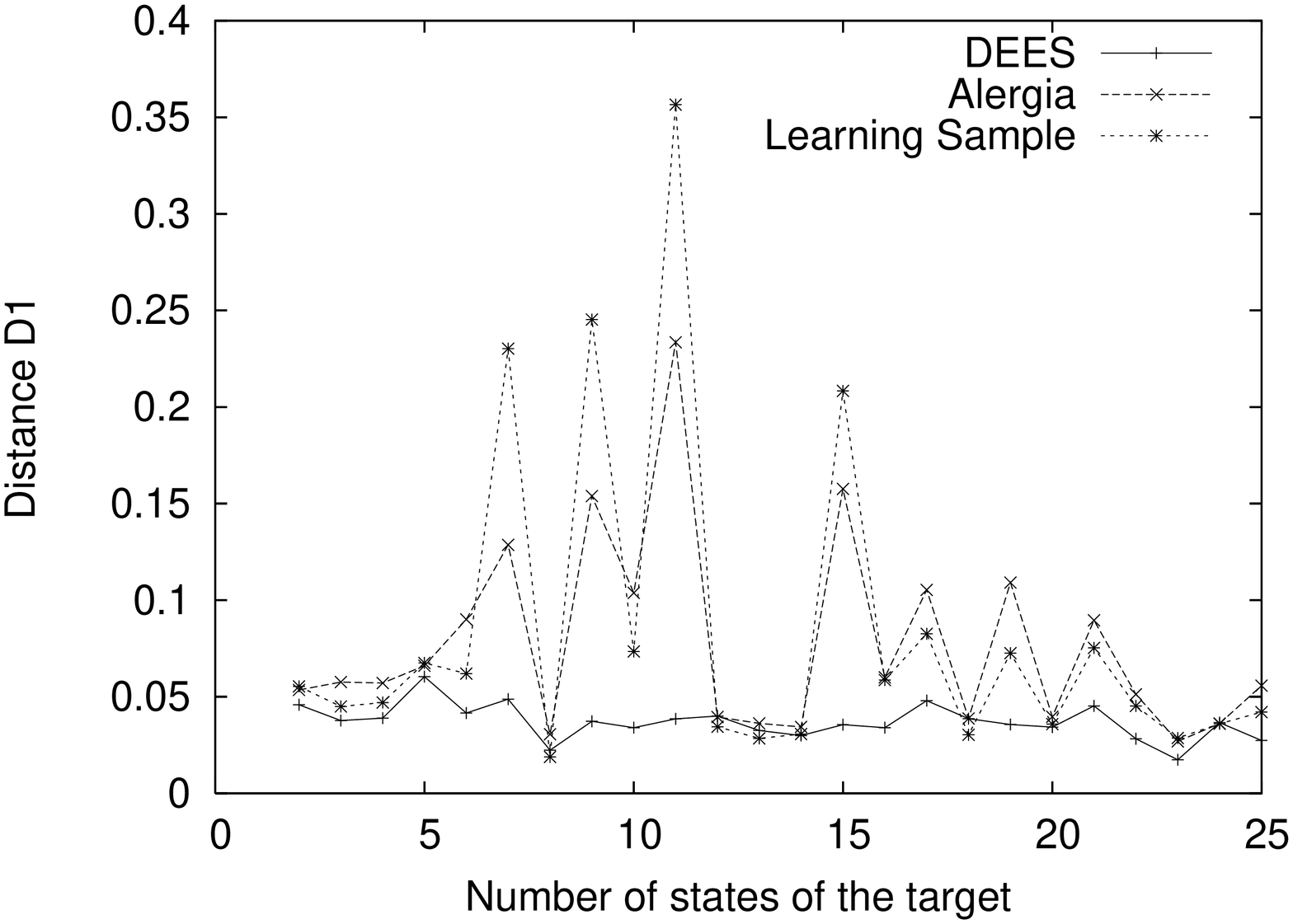}}& 
\subfigure[Size of the model.]{\includegraphics[angle=270,width=5.5cm]{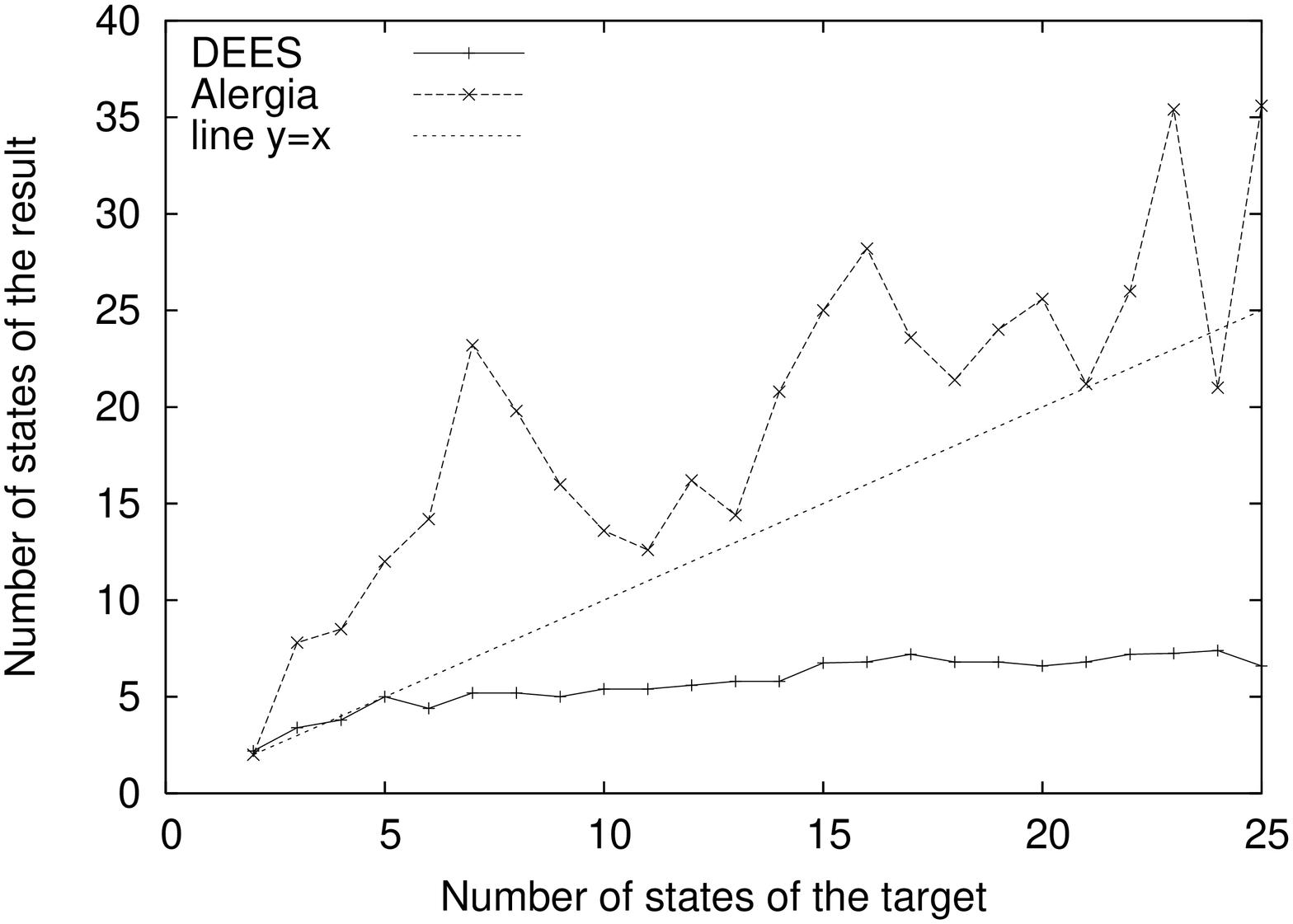}} 
\end{tabular}
\end{center}\vspace{-.9cm}
\caption{Results obtained from a set of PA generated randomly.~\label{RPA}}\vspace{-.6cm}
\end{figure}

We made another experiment in the class of PA. We study the
behavior of DEES when the learning samples are generated from
different targets randomly generated. For this experiment,  we take an alphabet of three
letters and we generate randomly some PA with a number of states from 2 to 25.  The PA are
generated in order to have a prefix representation which 
guarantees that all the states are reachable. The rest of the
transitions and the values of the parameters are chosen randomly.
Then, for each target, we generate 5 samples of size 300 times the
number of states of the target. We made this choice because we think
that for small targets the samples may be sufficient to find a good
approximation, while for bigger targets
there is a clear lack of examples. This last point allows us to see the
behaviors of the algorithms with small amounts of data. We learn an
automaton from each sample and compare it to the corresponding
target. Note that we didn't use MDI in this experiment because this
algorithm is extremely hard to tune, which implies an important cost in time
 for finding a good parameter.   The parameter of Alergia is fixed to a reasonable
value kept for all the experiment.
Results for Alergia and DEES are reported on Figure~\ref{RPA}. We also
add the empirical distance of the samples to the target automaton.  If
you consider the $D1$ norm, the performances of Alergia
depend highly on the empirical distribution.  Alergia infers
models close, or better, than those produced by DEES only when the empirical distribution is already very good, thus when it is not necessary to learn.
 Moreover, Alergia has a greater variance which implies a weak
robustness. On the other hand, DEES is always able to learn
significantly small models almost always better, even with small samples.

\begin{figure}[t]
\begin{center}
\begin{tabular}{ll}
\subfigure[Results with distance
$D1$]{\includegraphics[angle=270,width=5.5cm]{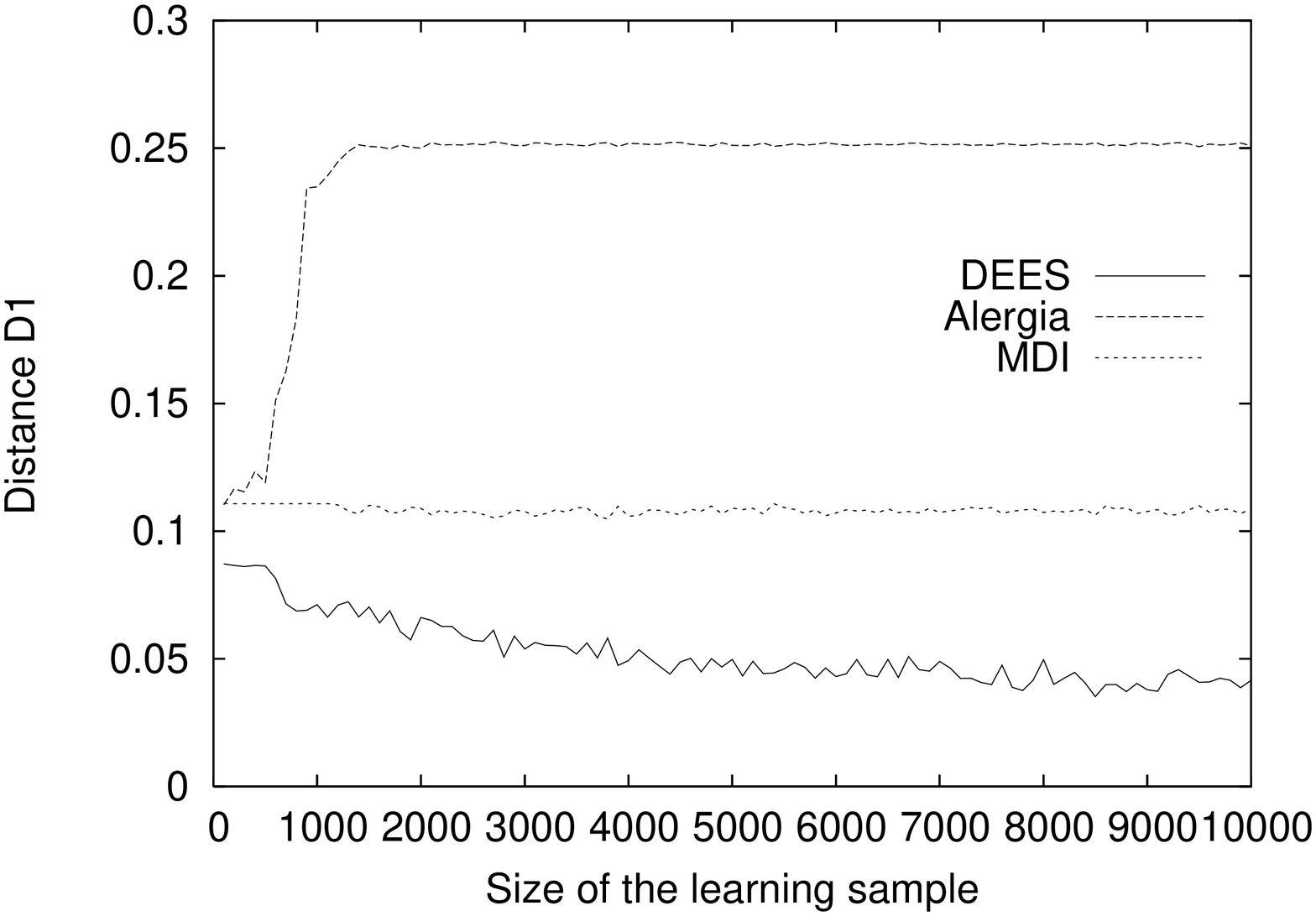}}
& 
\subfigure[Size of the model.]{\includegraphics[angle=270,width=5.5cm]{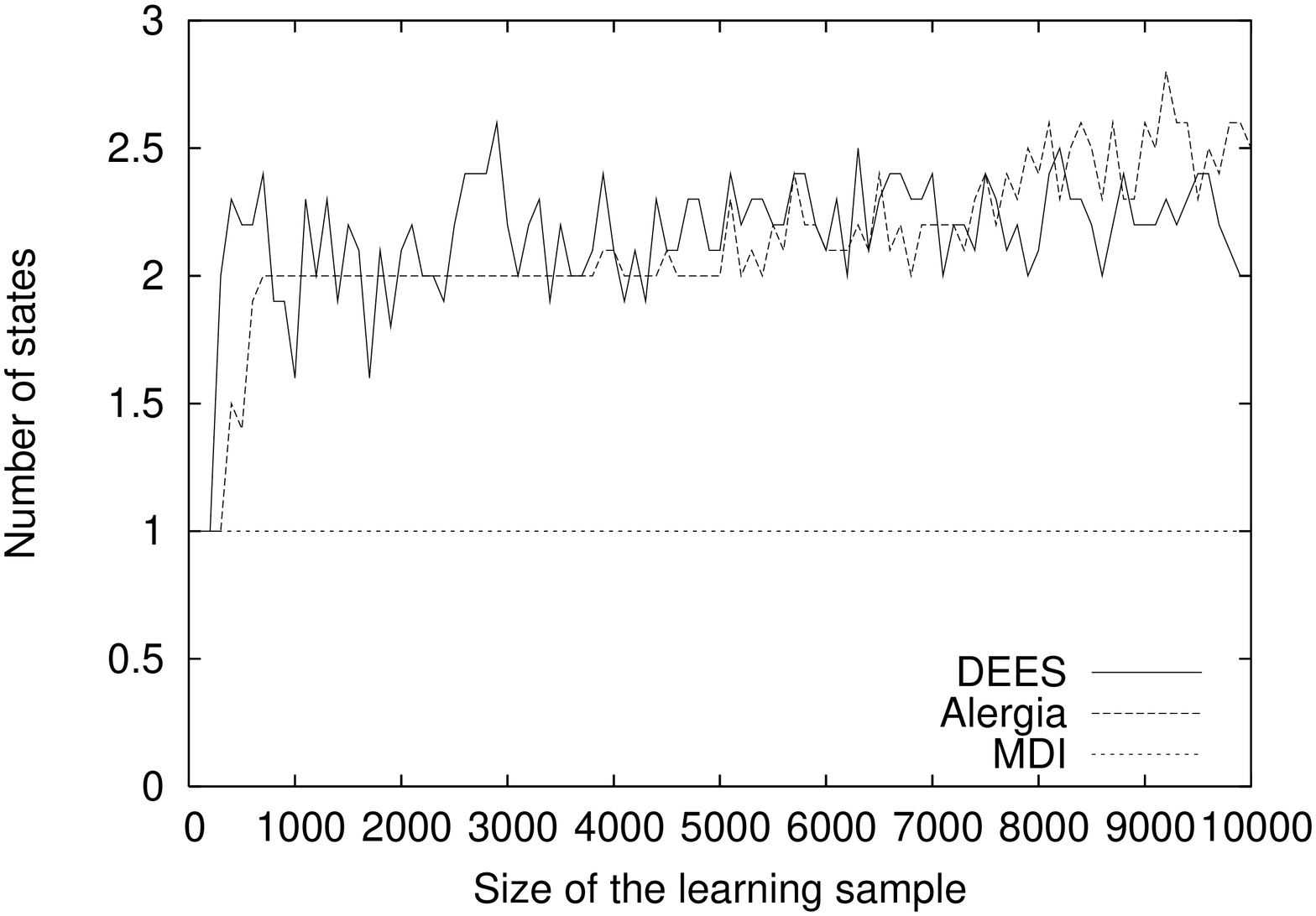}} 
\end{tabular}
\end{center}\vspace{-.9cm}
\caption{Results obtained with samples generated from a non rational
  stochastic language.\label{ExpPr}}\vspace{-.6cm}
\end{figure}

Finally, we carried out a last experiment where the objective is to
study the behavior of the three algorithms with samples generated from a
non rational stochastic language. We consider, as a target, the
stochastic language generated using the $p_r$ algorithm from the
automaton of Figure~\ref{fig:pseudoratnonrat} (note that this automaton admits a
prefix reduced representation of 2 states). We took $\rho=3/10$,
$\alpha=3/2$ and 
$\beta=5/4$. We follow the same experimental setup than the first experiment. Since we use rational
representations, we measure the distance
$D1$ from the automaton of Figure~\ref{fig:pseudoratnonrat} using a sample generated
by $p_r$ ({\it i.e.} we measure the $D1$ only for strings with a strictly positive value).
The results are
presented on Figure~\ref{ExpPr}. MDI and Alergia
are clearly not able to build a good estimation of the target
distribution and we  see that
their best answer is to produce a unigram. On the other hand, DEES is able to identify a
structure close to the MA that was used for defining the
distribution and produces good automata after 2000
examples. This means that DEES seems able to produce
pseudo-stochastic rational languages which are closed to a non
rational stochastic distribution. 

\section{Conclusion}
In this paper, we studied the class of pseudo-stochastic rational
languages (PSRL)  that are stochastic languages defined by multiplicity
automata which do not define stochastic languages but share some
properties with them. We showed that it is possible to decide wether an MA defines a PSRL, but we
can't decide wether two MA define the same PSRL. Moreover, it is possible
to define a
stochastic language from these MA but this language is not rational
in general. Despite of these drawbacks,
we showed experimentally that DEES produces MA computing
pseudo-stochastic rational languages
that provide good estimates of a target stochastic language. We
recall here that DEES
is able to output automata with a minimal number of parameters which
is clearly an advantage from a machine learning standpoint, especially
for dealing with small datasets. Moreover, our
experiments showed that DEES outperforms  standard probabilistic
grammatical inference approaches.  
Thus, we think that the class of pseudo-stochastic rational languages
is promising for many applications in grammatical inference. Beyond
the fact to continue the
 study of this class, we also plan to consider methods that could infer a class
 of MA strictly greater than the class of PSRL. We also began to work
 on an adaptation of the approaches presented in this paper to trees.

\bibliographystyle{plain}
\bibliography{report}
\newpage
\section{Annex}
\subsection{Proof of Lemma~\ref{lemmaRHO}\label{AnnexlemmaRHO}}
\setcounter{lemma}{0}
\begin{lemma}
Let $A$ be a reduced representation of a
pseudo-stochastic language. Then, $\rho(A)<1$.
\end{lemma}

\begin{proof}[sketch]
  Let $A=\left\langle \Sigma,Q,\varphi,\iota,\tau\right\rangle $ be
  a reduced representation of $r$ and let
  $B=\left\langle \Sigma,Q_B,\varphi_B,\iota_B,\tau_B\right\rangle $
  be an MA that computes $r$ and such that $\rho(B)<1$.  Since $A$ is
  reduced, the vector subspace $E$ of ${\mathbb R}\langle\langle \Sigma
  \rangle\rangle$ spanned by $\{r_{A,q}|q\in Q_A\}$ is equal to
  $[\{\dot{u}r|u\in\Sigma^*\}]$ and is contained in the vector
  subspace $F$ spanned by $\{r_{B,q}|q\in Q_B\}$. 

The set $\{r_{A,q}|q\in Q_A\}$ is a basis of $E$. Let us complete it
into a basis of $F$ and let $P_E$ be the corresponding projection
defined from $F$ over $E$. Note that for any $x\in \Sigma$ and any
$r\in F$, we have $P_E(\dot{x}r)=\dot{x}P_E(r)$.

For any state $q\in Q_B$, let us express $P_E(r_{B,q})$ in this
basis. $$P_E(r_{B,q})=\sum_{q'\in Q_A}\lambda_{q,q'}r_{A,q'}.$$

Note that for any MA $C$ and any state $q$ of $C$, $$\sum_{x\in
  \Sigma}\dot{x}r_{C,q}=\sum_{q'\in
  Q_C}\varphi_C(q,\Sigma,q')r_{C,q'}.$$

Therefore, for any state $q$ of $B$, we have $$P_E(\sum_{x\in
  \Sigma}\dot{x}r_{B,q})=P_E(\sum_{q'\in
  Q_B}\varphi_B(q,\Sigma,q')r_{B,q'})=\sum_{q'\in
  Q_B}\varphi_B(q,\Sigma,q')\sum_{q''\in Q_A}\lambda_{q',q''}r_{A,q''}$$
but also
 \begin{align*}
P_E(\sum_{x\in \Sigma}\dot{x}r_{B,q})=\sum_{x\in
  \Sigma}\dot{x}P_E(r_{B,q})&=\sum_{x\in \Sigma}\dot{x}\sum_{q'\in
  Q_A}\lambda_{q,q'}r_{A,q'}\\&=\sum_{q'\in
  Q_A}\lambda_{q,q'}\sum_{q''\in
  Q_A}\varphi_A(q',\Sigma,q'')r_{A,q''}\end{align*}
and therefore $$\sum_{q'\in
  Q_B}\sum_{q''\in
  Q_A}\varphi_B(q,\Sigma,q')\lambda_{q',q''}=\sum_{q'\in
  Q_A}\sum_{q''\in Q_A}\lambda_{q,q'}\varphi_A(q',\Sigma,q'').$$
Now,
let $M_A$ (resp. $M_B$, resp. $\Lambda$) be the matrix indexed by
$Q_A\times Q_A$ (resp. $Q_B\times Q_B$, resp. $Q_B\times Q_A$) and
defined by $M_A[q,q']=\varphi_A(q,\Sigma,q')$ (resp.
$M_B[q,q']=\varphi_B(q,\Sigma,q')$, resp.
$\Lambda[q,q']=\lambda_{q,q'}$). Note that the rank of $\Lambda$ is
equal to the dimension of $E$. We have $$M_B\Lambda=\Lambda M_A.$$

Let $\mu$ be an eigenvalue of $M_A$ and let $X$ an associated
eigenvector. We have $$M_B\Lambda X=\Lambda M_A =\mu\Lambda X$$
and
since the rank of $\Lambda$ is maximal, $\mu$ is also an eigenvalue of
$M_B$. Therefore, $\rho(B)<1$ implies that $\rho(A)<1$.\qed
\end{proof}

\subsection{Proof of Proposition~\ref{prop2MA}\label{Annexprop2MA}}
\setcounter{proposition}{2}
\begin{proposition}
It is undecidable whether two rational series define the same
stochastic language.
\end{proposition}

\begin{proof}
 Let $A=\langle \Sigma, Q, \iota, \varphi, \tau\rangle$ be an MA
  which satisfies $|r_A(u)|\leq \lambda^{|u|}$ for some
  $\lambda<1/(2|\Sigma|)$. Let $\overline{\Sigma}=\{\overline{x}|x\in 
  \Sigma\}$ be a disjoint copy of $\Sigma$ and let $c$ be a new
  letter: $c\not\in \Sigma\cup\overline{\Sigma}$. Let $u\rightarrow
  \overline{u}$ be the morphism inductively defined from $\Sigma^*$ into
  $\overline{\Sigma}^*$ by
  $\overline{\epsilon}=\epsilon$ and
  $\overline{ux}=\overline{u}\cdot\overline{x}$.
  
  Let $B=\langle \Sigma_B, Q, \iota, \varphi_B, \tau\rangle$ defined
  by $\Sigma_B=\Sigma\cup\overline{\Sigma}\cup\{c\}$,
  $\varphi_B(q,c,q')=1$ if $q=q'$ and 0 otherwise,
  $\varphi_B(q,x,q')=\varphi_B(q,\overline{x},q')=\varphi(q,x,q')$
  if $x\in \Sigma$. 
  
  Let $f$ be the rational series defined by $f(w)=r_A(uv)$ if
  $w=\overline{u}cv$ for some $u,v\in\Sigma^*$ and 0 otherwise.

Let $\rho$ be such that $2\lambda<\rho<1/|\Sigma|$, let $r$ be the
rational series defined on $\Sigma_B$ by $r(w)=\rho^{|w|}$ if $w\in
\overline{\Sigma}^*$ and 0 otherwise.  Let $g=f+r$. Check that 
$$\sum_{w\in\overline{\Sigma}^*}\rho^{|w|}=\sum_{n\geq
0}(|\Sigma|\rho)^n=\frac{1}{1-|\Sigma|\rho}\textrm{ and }$$
$$\sum_{u,v\in\Sigma^*}|f(\overline{u}cv)|=\sum_{u,v\in\Sigma^*}|r_A(uv)|\leq
\sum_{u,v\in\Sigma^*}\lambda^{|uv|}=\left(\sum_{n\geq 0}(|\Sigma|\lambda)^n\right)^2=\left(\frac{1}{1-|\Sigma|\lambda}\right)^2.$$

Therefore, the sum $\sum_{w\in \Sigma_B^*}g(w)$ is absolutely
convergent. Check also that {\small $$\sum_{w\in \Sigma_B^*}g(w)\geq \sum_{w\in
  \Sigma^*}\rho^{|w|}-\sum_{u,v\in \Sigma^*}|f(ucv)|\geq
\frac{1}{1-|\Sigma|\rho} -
\left(\frac{1}{1-|\Sigma|\lambda}\right)^2=\frac{|\Sigma|(|\Sigma\lambda^2-2\lambda+\rho)}{(1-|\Sigma|\rho)(1-|\Sigma|\lambda)^2}>0.$$}

Let $\mu=(\sum_{w\in \Sigma_B^*}g(w))^{-1}$ and $h=\mu g$.

For any $u\in \Sigma^*$, $h(\overline{u})=\mu\rho^{|u|}$,
$h(\overline{u}c\Sigma_B^*)=h(\overline{u}c\Sigma^*)=\mu
r_A(u\Sigma^*)$ and $h(\overline{u}\Sigma_B^*)=h(\overline{u}\overline{\Sigma}^*)+h(\overline{u}c\Sigma^*)=\mu(\frac{\rho^{|u|}}{1-|\Sigma|\rho^{|u|}}+r_A(u\Sigma^*))$.

Check also that for any $u\in \Sigma^*$,
$$\frac{\rho^{|u|}}{1-|\Sigma|\rho^{|u|}}+r_A(u\Sigma^*)\geq
\frac{\rho^{|u|}}{1-|\Sigma|\rho^{|u|}}-\sum_{v\in\Sigma^*}|r_A(uv|\geq
\frac{\rho^{|u|}}{1-|\Sigma|\rho^{|u|}}-\frac{\lambda^{|u|}}{1-|\Sigma|\lambda^{|u|}}>0.$$
Therefore,
$h(\overline{u})>0$ and $h(\overline{u}\overline{x}\Sigma_B^*)>0$ for
every $u\in \Sigma^*$ and any letter $x\in \Sigma$. On the other hand,
$h(\overline{u}c\Sigma_B^*)>0$ iff $r_A(u\Sigma^*)\leq 0$. That is,
$p_h=p_r$ iff $r_A(u\Sigma^*)\leq 0$ for every $u\in \Sigma^*$. An
algorithm capable to decide whether $p_h=p_r$ could be used to decide
whether $r_A(u\Sigma^*)\leq 0$ for every $u\in \Sigma^*$. \qed
\end{proof}

\subsection{Drawing a word according to $p_r$ \label{APR}}
Modification of Algorithm~\ref{A:PR} in order to draw a word according
to the distribution $p_r$.
\begin{algorithm}[t]
\KwIn{ an MA
    $A=\left\langle \Sigma,Q=\{q_1,
    \ldots,q_n\},\varphi,\iota,\tau\right\rangle $ s.t. $\rho(A)<1$ and
     $r_A(\Sigma^*)=1$ }
\KwOut{ a word $u$ drawn according to $p_{r_A}$}
\For{$i=1,\ldots,n$ /* this step is polynomial in $n$ and is done
  once*/}{
$s_i\leftarrow r_{A,q_i}(\Sigma^*)$; \hspace{.5cm} $e_i\leftarrow \iota(q_i)$\;
}
$u\leftarrow \varepsilon$\;
$finished \leftarrow false$\;
$w\leftarrow \varepsilon$; \hspace{.5cm} $\lambda \leftarrow 1$ /* $\lambda$ is
equal to $p_{r_A}(w\Sigma^*)$*/ \;
\While{not $finished$}{
$S\leftarrow \emptyset$\;
$\lambda \leftarrow 1$ \;
 $v \leftarrow \sum_{i=1}^n e_i\tau(q_i)$\;
\lIf{$v>0$}{$S\leftarrow  \{(\varepsilon,v)\}$\;$\lambda\leftarrow v$\;}
\For{$x\in \Sigma$}{
$v \leftarrow \sum_{i,j=1}^n e_i\varphi(q_i,x,q_j)s_j$\;
\If{$v>0$}{$S\leftarrow S \cup \{(x,v)\}$\;$\lambda\leftarrow\lambda+v$\;}}
\lFor{$(x,v)\in S$}{$(x,v) \leftarrow (x,v/\lambda)$\;}
$x\leftarrow Draw(S)$ /*Draw randomly an element $(x,p)$ of $S$ with
probability 
$p$*/\;
\eIf{$x=\varepsilon$}{$finished\leftarrow True$\;}{$u\leftarrow
  ux$\;\lFor{$i=1,\ldots,n$}{$e_i\leftarrow \sum_{j=1}^n e_j\varphi(q_j,x,q_i)$}\;}
}
\caption{Algorithm drawing a word according to the distribution $p_r$.\label{A:PR2}}
\end{algorithm}

\end{document}